\pdfoutput=1

\documentclass[11pt]{article}

\usepackage[preprint]{acl}

\usepackage{times}
\usepackage{latexsym}

\usepackage[T1]{fontenc}

\usepackage[utf8]{inputenc}

\usepackage{microtype}

\usepackage{inconsolata}

\usepackage{graphicx}

\usepackage{gal_basic}

\usepackage{xspace}
\usepackage{xcolor}
\usepackage[inline]{enumitem}
\usepackage{subfig}
\usepackage{tabularx}
\usepackage{booktabs}
\usepackage{amssymb}
\usepackage{makecell}

\newcommand\modelsampling{\textsc{ModelAnswers}\xspace}
\newcommand\baseline{\textsc{TrainCounts}\xspace}
\newcommand\mixture{\textsc{Mixture(M)}\xspace}
\newcommand\genericmixture{\textsc{Mixture}\xspace}
\newcommand\uniformbaseline {\textsc{UniformIntegers}\xspace}
\title{Keep Guessing? When Considering Inference Scaling, Mind the Baselines}

\author{First Author \\
  Affiliation / Address line 1 \\
  Affiliation / Address line 2 \\
  Affiliation / Address line 3 \\
  \texttt{email@domain} \\\And
  Second Author \\
  Affiliation / Address line 1 \\
  Affiliation / Address line 2 \\
  Affiliation / Address line 3 \\
  \texttt{email@domain} \\}

\author{Gal Yona$^{*\gamma}$ \quad Or Honovich$^{*\tau\gamma}$ \quad Omer Levy$^{\delta}$ \quad Roee Aharoni$^\gamma$ \\ [10px]
    $^\tau$Tel Aviv University \quad
    $^\gamma$Google Research \quad  $^\delta$Google DeepMind  \quad  \\
}

\begin{document}
\maketitle
\begin{abstract}
Scaling inference compute in large language models (LLMs) through repeated sampling consistently increases the coverage (fraction of problems solved) as the number of samples increases.
We conjecture that this observed improvement is partially due to the answer distribution of standard evaluation benchmarks, which is skewed towards a relatively small set of common answers. To test this conjecture, we define a baseline that enumerates answers according to their prevalence in the training set.
Experiments spanning two domains -- mathematical reasoning and factual knowledge -- reveal that this baseline outperforms repeated model sampling for some LLMs, while the coverage for others is on par with that of a mixture strategy that obtains $k$ answers by using only $10$ model samples and similarly guessing the remaining $k-10$ attempts via enumeration.  
Our baseline enables a more accurate measurement of how much repeated sampling improves coverage in such settings beyond problem-agnostic guessing.

{\let\thefootnote\relax\footnotetext[0]{{$^*$ Equal Contribution, order determined at random; Correspondence: \tt{roeeaharoni@google.com}.}}}

\end{abstract}
\begin{figure}[t]
\setlength{\belowcaptionskip}{-10pt}
    \centering
    \includegraphics[width=0.85\linewidth]{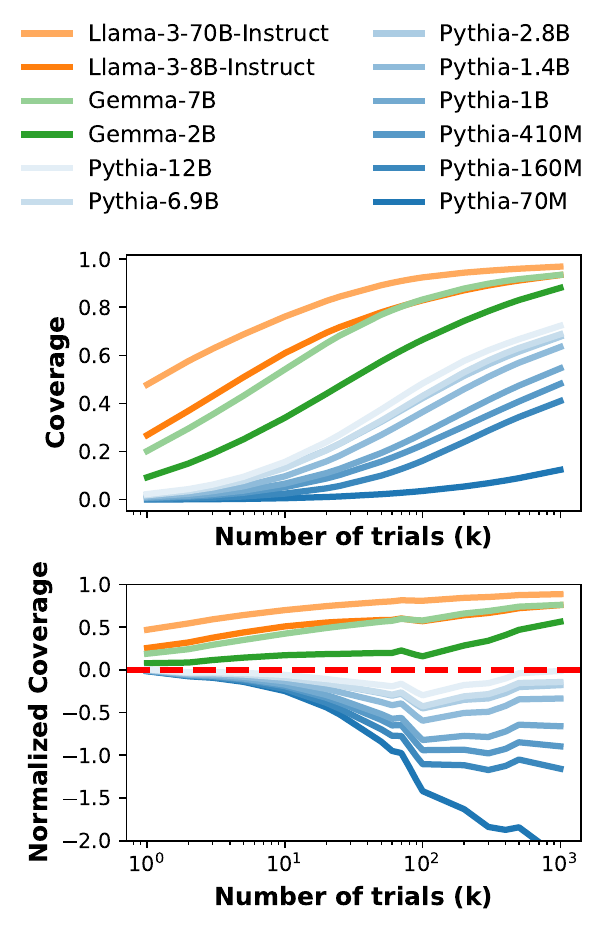}
    \caption{Standard coverage curves  (top; see Fig. 3 in \citet{brown2024largelanguagemonkeys}) vs  normalized coverage curves (bottom), for the MATH dataset. Normalized coverage is obtained by re-scaling the improvements relative to our \baseline baseline. 
    We see that despite all Pythia models (\textbf{\textcolor{mylightblue}{blue}}) showing non-negligible coverage gains, these are actually worse than simple answer enumeration (below the $y=0$ dashed \textbf{\textcolor{red}{red}} line).
    }
    \label{fig:normalized_coverage_math}
\end{figure}

\section{Introduction}
Scaling \emph{training} compute -- larger models, larger datasets, and longer training runs -- has been a main driver of progress in LLMs \cite{kaplan2020scalinglawsneurallanguage, hoffmann2022scaling}. Recent works  highlighted the benefit of additionally scaling \emph{inference} compute: sampling longer sequences, e.g., chain-out-thought sampling \cite{wei2022chain, star2022}, using increasingly longer input contexts \cite{levy-etal-2024-task}, and repeatedly sampling model responses \cite{brown2024largelanguagemonkeys, hassid2024larger}.
In particular, repeated sampling increases the fraction of problems solved by at least one of $k$ attempts (known as \emph{coverage} or pass@$k$) as $k$ grows, as demonstrated across a variety of tasks such as code generation and mathematical reasoning.
For verifiable tasks, these gains can be used to post-train models \cite{hosseini2024v}.

In this work, we argue that measuring
coverage alone overlooks the fact that some commonly used datasets have a closed, virtually enumerable answer set -- possibly making them easy-to-guess given enough attempts. This raises a fundamental question: could the observed coverage gains be partially attributed to \emph{lucky guesses}, rather than uncovering \emph{correct} reasoning? This observation has implications not only for model evaluation, but also for post-training, when utilizing model-sampled answers for self-improvement. This process relies on selecting chains with verified final answers out of multiple samples,
bearing the risk of inadvertently rewarding solutions that are ``right for the wrong reasons'', i.e., incorrect reasoning chains ending in a correct final answer.

To quantify how much repeated sampling improves beyond guessing, we establish simple baselines and report coverage gains as their relative improvement over these baselines.
Specifically, we compare three approaches:
(1) \modelsampling{}: Where $k$ candidate answers are obtained by sampling $k$ model responses.
(2) \baseline{}: A baseline where $k$ candidate answers are obtained by enumerating the $k$ most frequent answers in the training set. 
(3) \mixture{}: A mixture strategy, where the first $M$ answers are obtained by \modelsampling{}, and the remaining $k-M$ answers are obtained using \baseline{}. 

We experiment with mathematical reasoning \cite[MATH;][]{hendrycks2021measuring}, where solutions include reasoning chains, and factual knowledge \cite[EntityQuestions;][]{Entity_Questions}, which includes no chains. We find that:
\begin{itemize}
[itemsep=1pt, topsep=2pt,leftmargin=*]
\item Normalizing the coverage improvements compared to the baseline reveals that some models actually perform worse than problem-agnostic enumeration (see Figure \ref{fig:normalized_coverage_math});
\item Even for the models with high normalized coverage gains, we observe that \mixture{} with small values of $M$ achieves coverage nearly as good as \modelsampling{} (at a small fraction of the cost), suggesting that models either ``know'' the correct answer, or cannot do much better than informed guessing.
\end{itemize}

Our findings suggest that some commonly used datasets become degenerate when considering large-scale repeated sampling. 
While inference scaling seems like a promising approach for improving performance, we suggest carefully selecting datasets, models, and baselines when assessing this method, and interpreting results with caution.
\section{Repeated Sampling}\label{sec:background}

Recent efforts have scaled inference compute by performing repeated sampling with thousands of samples, focusing on tasks where candidate solutions are evaluated as either right or wrong.
Repeated sampling is usually evaluated via (1) \emph{coverage}, i.e., the fraction of problems that can be solved correctly by at least one of the sampled model answers, and (2) \emph{precision}, i.e., the ability to identify the correct answer from a set of sampled answers. For tasks with automatic verification (e.g., unit tests for coding) an increase in coverage can translate to model improvements \cite{hosseini2024v}.

Recent work found striking coverage gains by scaling the number of sampled answers. \citet{brown2024largelanguagemonkeys} showed that while the Pythia-160M model solves only $0.27\%$ of the problems in MATH with a single attempt, the coverage using $k=10,000$ attempts reaches as far as $57\%$. 
Similarly, \citet{hassid2024larger} showed that for code generation, repeated use of smaller models yields consistent improvements, with gains of up to $15\%$ given automatic verification.
A possible interpretation of these results is that even very small models are more capable than previously assumed, such that repeated sampling combined with a strong verifier may unlock this seemingly ``hidden potential.'' 
\section{Baselines for Repeated Sampling}\label{sec:baselines}

We focus on coverage gains due to repeated model sampling, and argue that their proper interpretation requires comparing them against the gains of simple ``guessing'' strategies. We thus
compare the standard notion of coverage (\S\ref{sec:model_answers}) to two simple baselines: answer enumeration based on answer counts in the training data (\S\ref{sec:train_counts}) and a mixture strategy that combines answer enumeration with few model samples (\S\ref{sec:mixture}). 

\subsection{\modelsampling{}: Repeated sampling}
\label{sec:model_answers}

\modelsampling{} is the standard repeated model sampling. Here, coverage (pass@$k$) is defined as the expected number of problems that are solved by at least one model answer, when sampling $k$ answers. Following prior work \cite{chen2021evaluating}, we estimate pass@$k$ for each problem $i$ by sampling $N=1000$ samples and using the unbiased estimator $1 - \frac{\binom{N - C_i}{k}}{\binom{N}{k}} $, where $C_i$ is the number of correct samples for problem $i$.

\subsection{\baseline{}: Answer Enumeration} 
\label{sec:train_counts}
Our na\"{i}ve guessing strategy, \baseline{}, relies on obtaining the answer counts in the training data of the dataset under consideration. Here, pass@$k$ is the fraction of problems\footnote{While the standard pass@$k$ is a random variable depending on the sampled answers, pass@$k$ for our guessing strategy is a deterministic quantity depending only on the (fixed) guesses and the ground truth labels.} for which the correct answer is among the top-$k$ most frequent answers in the training set. As this strategy relies on the training-set answer distribution, we refer to it as ``informed enumeration''.
Note that \baseline{} is a problem-agnostic strategy, which predicts the same $k$ answers, regardless of the tested input problem.

We stress that we do not suggest using \baseline{} as a prediction method, but solely use it to critically examine repeated model sampling.

\subsection{\genericmixture{}: First Sample, Then Guess}
\label{sec:mixture}

We additionally consider a \emph{mixture} strategy that combines both model samples and enumerated answers. Specifically, for \mixture, we obtain $M$ answers by sampling from the model, while the remaining $k-M$ answers are obtained using the most frequent answers in the training set (as in \S\ref{sec:train_counts}). We estimate pass@$k$ as the fraction of problems for which the correct answer is among the $M$ randomly selected model answers or the top $k-M$ answers in the training set (averaged over $T$ random draws from a given set of $1000$ sampled model answers). We use $M$ values of 1, 5, and 10.

\section{Experimental Setup}

\paragraph{Datasets.} We focus on two domains: mathematical reasoning and factual knowledge.
\begin{itemize}
[itemsep=1pt, topsep=2pt,leftmargin=*]
\item \textbf{MATH}: A dataset of challenging math word problems \cite{hendrycks2021measuring}. We use the same 128 randomly selected test problems used in \citet{brown2024largelanguagemonkeys}.
\item \textbf{Entity Questions (EQ)}: A QA dataset \cite{Entity_Questions} with diverse questions about various entities. We sample 128 questions, while maintaining a balanced proportion of relations. 
\end{itemize}

\paragraph{Models.}
For MATH, we use the data from \citet{brown2024largelanguagemonkeys}, containing three model families: Llama \cite{llama3model}, Gemma \cite{gemma1} and Pythia \cite{biderman2023pythia}. For EQ, we use models from the Gemma 2 \cite{gemmateam2024gemma2improvingopen} and Gemini \cite{team2023gemini} model families. See the full list of models in Appendix~\ref{sec:appendix_models}.

\paragraph{Obtaining Training Set Counts.} In MATH, we obtain answer counts using the entire train split. In EQ, when guessing an answer to a question from a relation $r$, we select an answer according to counts obtained only from the train set of $r$. See answer statistics for MATH in Table~\ref{table:math_counts} in the appendix.

\paragraph{Answer Verification.} Measuring coverage requires verifying candidate answers for correctness.
For MATH, we apply the evaluation protocol of \citet{brown2024largelanguagemonkeys}, and for EQ, we use an F1-based evaluation protocol (details in appendix~\ref{sec:appendix_answer_verification}).

\section{Results}\label{sec:results}

\begin{figure*}[t]
    \centering
    \includegraphics[width=1.0\linewidth]{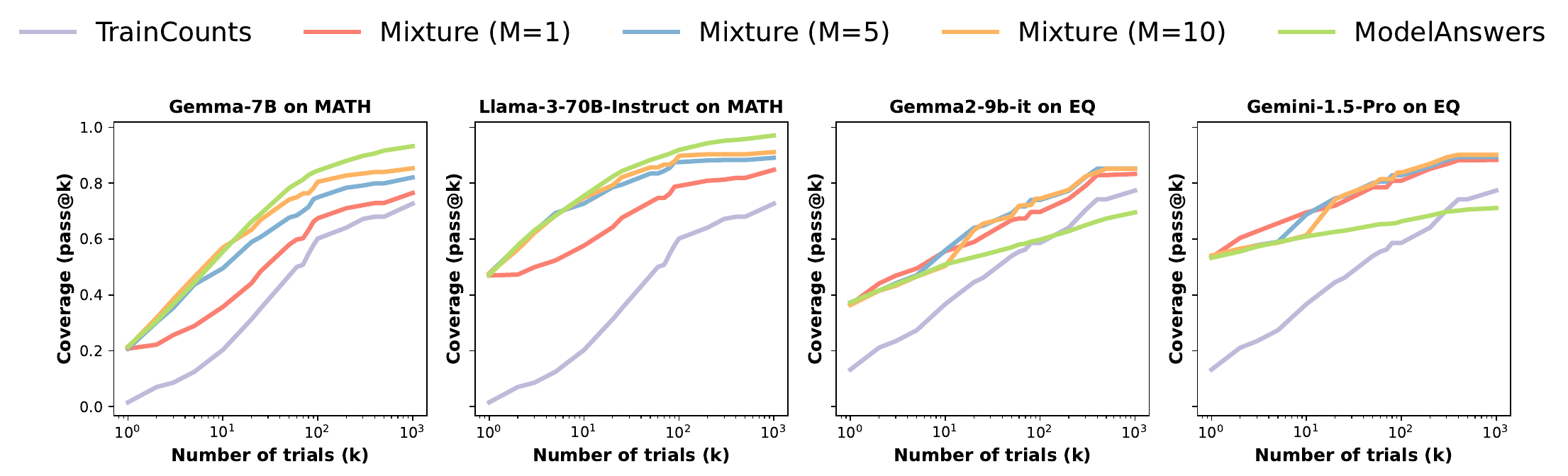}
    \caption{Coverage curves for model sampling and guessing-based answers for MATH (Llama-3-70B and Gemma-7B models) and EQ (Gemini-1.5-Pro and Gemma 2-9b models). We evaluate \mixture{} for $M=1,5,10$ and compare \mixture{} with \baseline{} and \modelsampling{}.}
    \label{fig:mixture}
    \vspace{-0.2cm}
\end{figure*}



\paragraph{Baselines Can Outperform Thousands of Model Samples.}
To compare coverage gains obtained from model sampling to guessing-based gains,
we calculate a normalized notion of pass@$k$ that quantifies the added gain of using \modelsampling{} compared to \baseline{}:
\begin{equation*}
\tiny
    \frac{\text{Coverage(\modelsampling{}) - Coverage(\baseline{})}}{ 1 - \text{Coverage(\baseline{})}}
\end{equation*}
The results for MATH are in Figure~\ref{fig:normalized_coverage_math}. The results for EQ, showing similar trends, are detailed in Figure~\ref{fig:normalized_coverage_eq} in the appendix.

While the unnormalized curves demonstrate significant increases in coverage across all models, explicitly quantifying the gains over an enumeration baseline shows notably smaller gains, with some models (e.g. all Pythia models) performing worse than the baseline. 

This highlights that repeated-sampling results should be interpreted with caution, especially for datasets that were not originally designed for scenarios of thousands of solution attempts.

\paragraph{Are Few Samples All You Need?}
We speculate that models either ``know'' the correct answer or cannot do much better than guessing. To test this, we compare \modelsampling{} with \mixture{}.
Results for two models on both MATH and EQ are shown in Figure~\ref{fig:mixture}, and for additional models in Figure~\ref{fig:mixture_additional} in the appendix.  All results are inline with our hypothesis: while for MATH we observe a considerable gap between \baseline{} and \modelsampling{}, \mixture{} closes most of this gap, even for small values of $M$. E.g., for Llama-3-70B, the pass@$k$ values at $k=1000$ are 97\%, 91\% and 73\% for \modelsampling{}, \mixture{} with $M= 10$, and \baseline{} respectively. We conclude that sampling as few as 10 model answers and proceeding with guesses yields similar gains as $k$ model samples, though at a significantly lower compute budget. These results are consistent across all $k$ values.

The results for EQ show a different trend: for large-enough $k$ values, \baseline{} even outperforms \modelsampling{}, with \mixture{} outperforming both \baseline{} and \modelsampling{}. These results are due to the fact that some relations in EQ are highly challenging for the tested models, while having a rather small set of possible answers and a large fraction of the test set answers present in the training set -- all making answer enumeration a strong baseline. \mixture{} results per relation are shown in Appendix~\ref{sec:eq_breakdown} .

Together, our results show that the overwhelming benefit of repeated sampling is materialized within the first few samples, casting doubts on whether scaling the number of samples by several orders of magnitude is necessary in these settings.

\begin{table}[t]
    \small
    \centering
    \begin{tabularx}{\linewidth}{@{}X@{}}
    \toprule
     \makecell[X]{\textbf{Question}: Josh and Mike live 13 miles apart. Yesterday Josh started to ride his bicycle toward Mike's house. A little later Mike started to ride his bicycle toward Josh's house. When they met, Josh had ridden for twice the length of time as Mike and at four-fifths of Mike's rate. How many miles had Mike ridden when they met?\\ \vspace{-5pt} \textbf{CoT:} If Josh had ridden for twice the length of time as Mike, then Josh had ridden for a total of $2\cdot5=10$ miles. If Mike had ridden for $x$ miles, then Mike had ridden for a total of $4x$ miles. Subtracting, we have $10=4x$. Therefore, $x=5$. Final Answer: The final answer is $5$ miles. I hope it is correct.} \\
    \bottomrule
    \end{tabularx}
    \caption{A Pythia-12B response with a correct final answer but a nonsensical CoT. We observe this behavior for 9/10 inspected examples. All 10 problems and model generations can be found \href{https://docs.google.com/spreadsheets/d/1SWE5mOeKAdb1nKAk_jnC3Z2H0jol70CWjKicc4x7FZI/edit?usp=sharing}{here}. 
    }
    \label{tab:pythia-cot-main}
    \vspace{-0.25cm}
\end{table}

\section{Analysis and Discussion}
\label{sec:discussion}

\paragraph{Weaker than baseline means useless CoTs?}
We hypothesize that for models who do worse than our simple guessing baseline, reasoning chains that end in a correct answer may be incorrect when closely inspected \cite{turpin2024language}. To test this, we inspect solutions from the best model that under-performs our baseline (Pythia 12B). We sample 10 CoTs that have a correct final answer and find that in 9/10 cases, the reasoning chains are indeed incorrect, while the answer is correct; see Table~\ref{tab:pythia-cot-main}.  This highlights the potential risk of common self-improvement pipelines \cite{zelikman2022star, liu2023statistical}, that reward the entire CoT based only on the correctness of the final verdict.

\paragraph{Answer enumeration vs i.i.d sampling.}
Our random baseline differs from \modelsampling{} in that it uses enumeration over a fixed set of candidate solutions, rather than i.i.d sampling from a fixed distribution. 
Hence, pass@$k$ for repeated sampling typically relies on $k'\ll k$ unique answers, while for \baseline{}, it uses $k$ unique values. This design decision of our programmatic baselines aims to shed light on the usefulness of inference scaling, and to point at some of its shortcomings. Inference scaling approaches that would attempt to maximize answer diversity may outperform our baselines with higher margins.

\section{Conclusions}
\label{sec:conclusions}
We provide a critical perspective on the coverage gains obtained by repeatedly sampling model answers, showing that for weaker models, the gains are often worse than simple guessing baselines, and that for stronger models, much of the gains can be obtained with as few as 10 model samples. Properly accounting for the actual benefit of repeated sampling is an important and timely objective, given both the potential negative implications of rewarding incorrect CoTs and the computational costs associated with sampling thousands of model responses. Our findings shed light on how commonly used datasets can become degenerate when large-scale sampling is involved, reiterating the importance of using challenging benchmarks, where success by chance is unlikely.

\clearpage 
\newpage
\section{Limitations}

Our work critically examines the utility of inference scaling via repeated sampling. We show that when re-scaling the improvements relative to the performance of simple answer enumeration baselines, the gains are less pronounced, with some smaller models even performing worse than the baseline.

In our study, we used tasks that have overall structured outputs and demonstrated high performance of answer-counts-based baseline. For tasks that have free-form or longer outputs, this baseline is not directly applicable. However, we believe that most of our observations would still apply. Consider, for example, the question \nl{Which land mammal has the longest tail?} from the 
NaturalQuestions dataset \cite{kwiatkowski2019natural}. 
While the correct answer, \emph{giraffe}, is not prevalent in the training set, one would still be able to guess it correctly, given enough attempts (e.g., by enumerating the set of known land mammals).  Extending our guessing baselines to such datasets -- possibly drawing inspiration from how humans make informed guesses -- is an interesting direction for future exploration.

From a technical perspective, our experiments examine inference scaling by taking up to $k=1000$ samples (rather than $10,000$, as done by \citet{brown2024largelanguagemonkeys}). We do so due to efficiency considerations and stress that this does not affect our conclusions.



\bibliography{custom, anthology_0, anthology_1}

\clearpage
\appendix
\newpage
\section{Related Work}
\label{sec:related}

\paragraph{Inference Scaling.}
Utilizing additional computational resources during inference is carried out across different axis, such as generating more tokens before converging into a final answer \cite{wei2022chain, kojima2022large, zelikman2022star}, including increasingly longer input contexts \cite{shaham-etal-2022-scrolls, team2023gemini, bertsch2024incontext, levy-etal-2024-task}, or by sampling few model answers and selecting the most consistent one \cite{wang2023selfconsistency}.
Recently, there has been growing interest in large-scale model sampling - i.e., sampling orders of thousands of model answers. \citet{hassid2024larger} showed that for code generation, repeated use of smaller models yields consistent improvements, with gains of up to 15\% given automatic verification. \citet{brown2024largelanguagemonkeys} tested repeated sampling for code generation and mathematical reasoning, by measuring the coverage -- fraction of problems solved by any attempt - for different quantities of model samples, showing that the coverage scales log-linearly with the number of samples. Notably, they manually verified 105 chains of thought of 105 correct Llama-3-8B-Instruct predictions on GSM8K \cite{cobbe2021gsm8k}, finding that over 90\% of the chains are valid. We show that this is not the case for correct Pythia 12B predictions on MATH, suggesting that the observed coverage gains are more due to \nl{lucky guessing} than a result of correct but unlikely answers.

\paragraph{Self-Improvement.}

A common approach for improving the reasoning abilities of LLMs during post-training is self-improvement, which relies on updating a model based on solutions generated by the model itself \cite{zelikman2022star, liu2023statistical, gulcehre2023reinforced}. In self-improvement, several answers (CoT and final answer) are sampled from the model, while only \nl{correct} generations are rewarded. A candidate answer can be determined \nl{correct} using automatic verification (when applicable, e.g. unit tests in coding problems) or oracle labels (comparing the final answer to a ground truth answer). Since this recipe only considers the final answer as supervision, it may end up inadvertently rewarding answers that are “right for the wrong reasons” if such answers are generated by the model to begin with. Our approach provides simple ways to measure whether this behavior is likely by comparing the coverage improvements with those obtained by simple answer enumeration. Recent work has also considered employing intermediate rewards \cite{ni2022learning}, providing a finer-grained signal for intermediate steps within the CoT.  

\paragraph{Random Baselines.}

Reporting the results of simple baselines has an important role in machine learning, helping to contextualize performance \cite{lipton2019troubling}, diagnose dataset issues \cite{zheng2024cheating} and reveal possible shortcut solutions \cite{geirhos2020shortcut}. Such baselines include random baselines, majority (i.e., always predicting the most prevalent class), and simple heuristics. Random baselines are cleanly defined in classification tasks as the expected accuracy of guessing labels uniformly at random. Beyond classification tasks (e.g. natural language generation), however, the strategy itself is not clearly defined. The types of tasks we consider in this work can be approximately viewed as classification tasks, in the sense that the set of possible final answers is approximately enumerable. In the context of LLMs,  \cite{yauney2024stronger} recently proposed a stronger random baseline that accounts for scenarios of reusing the evaluation data.

\section{Experimental Setup}
\label{sec:appendix_experimental}
\subsection{Models}\label{sec:appendix_models}
For MATH, we use the data from \citet{brown2024largelanguagemonkeys}, spanning three model families:
\begin{itemize}
    \item \textbf{Llama 3}: Llama-3-8B, Llama-3-8B-Instruct, Llama-3-70B-Instruct \cite{llama3model}.
    \item \textbf{Gemma}: Gemma-2B, Gemma-7B \cite{gemma1}.
    \item \textbf{Pythia}: Pythia-70M through Pythia-12B (eight models in total) \cite{biderman2023pythia}.
\end{itemize}

For EQ, we use models from the Gemma 2 and Gemini model families:
\begin{itemize}
    \item \textbf{Gemini}: Gemini-Flash, Gemini-1.5-Pro \cite{team2023gemini}.
    \item \textbf{Gemma 2}: Gemma 2-2b-it, Gemma-2-9b-it \cite{gemmateam2024gemma2improvingopen}\footnote{We use the HuggingFace API for obtaining predictions. \\\url{https://huggingface.co/}}.
\end{itemize}

\subsection{Answer Verification}\label{sec:appendix_answer_verification}
Measuring the coverage requires verifying the correctness of each candidate answer, whether model-sampled or guess-based.
For MATH, we use an oracle verifier that checks if the candidate answer is mathematically equivalent to the correct answer, as in \citet{brown2024largelanguagemonkeys}. For EQ, as there is no oracle verifier available, we calculate a token-level F1 score and consider an answer as correct if the F1 score exceeds a threshold of $0.5$, as in \citet{yona-etal-2024-narrowing}. 
This verifier may miss some cases (e.g. due to rephrasing and granularity mismatches). However, manual inspection shows that this is a reliable metric.  
\subsection{Sampling from Entity Questions}

To derive answers for EntityQuestions, we use a similar procedure to the one used in \citet{brown2024largelanguagemonkeys}. We sample using a temperature of 0.7, taking 1000 predictions for each question. See Table 
 \ref{table:eq-sampling} for the full details.

\begin{figure}[ht]
\centering
\footnotesize
\begin{tabular}{|p{0.2\textwidth}|p{0.2\textwidth}|}
\hline
\textbf{Instruction Prompt} & \textbf{Example Few-Shot Demonstration} \\
\hline
Answer the given question. Provide your answer directly, without any prefixes.\newline\newline Here are some examples:
 &
Question: What position does Diego Rivero play?\newline\newline Answer: midfielder \\
\hline
\end{tabular}
\caption{Prompt used for sampling answers for EntityQuestions. To generate an answer for a target test question, we append to the instructions (left) five randomly selected training examples from the same relation (an example is shown on the right).}
\label{table:eq-sampling}
\end{figure}

\section{Data Statistics}

\begin{table}[htbp]
\centering
\resizebox{0.45\textwidth}{!}{ 
\small  
\begin{tabular}{cccccccccc}  
\hline
2  & 1  & 3  & 6  & 5  & 4  & 8  & 0  & 12 & 10 \\
9  & 7  & 16 & 15 & 20 & 11 & $\frac{1}{2}$ & 60 & 13 & 24 \\
18 & -1 & 30 & 14 & 17 & -2 & 36 & 25 & 32 & 28 \\
120 & 50 & 21 & $\frac{1}{3}$ & 40 & $\frac{1}{4}$ & 100 & 19 & 90 & 27 \\
$\frac{2}{3}$ & 35 & -3 & $\frac{3}{4}$ & 26 & 72 & 22 & 45 & -6 & -4 \\
150 & 23 & 64 & 31 & 48 & $\frac{3}{2}$ & 80 & 29 & 38 & $\frac{3}{5}$ \\
56 & 49 & 96 & 84 & 144 & 57 & 41 & 200 & $\frac{5}{2}$ & $\frac{1}{6}$ \\
$\frac{1}{8}$ & 81 & $\frac{4}{3}$ & 108 & 42 & 39 & 52 & 34 & $\sqrt{3}$ & 47 \\
$\frac{4}{5}$ & -5 & 70 & 54 & 63 & 59 & 33 & $\sqrt{2}$ & -$\frac{1}{2}$ & 400 \\
98 & 75 & 51 & 61 & 58 & 37 & 140 & 73 & 112 & -8 \\
\hline
\end{tabular}
}
\caption{The 100 most frequent answers in the training set of MATH dataset. Among the top-$k$ frequent training set answers, the fraction of answers that are integer solutions is 85\% (for $k=100$) and 48.9\% (for $k=1000$).}
\label{table:math_counts}

\end{table}

\section{Random Baseline}
\begin{figure}[t]
\setlength{\belowcaptionskip}{-8pt}
    \centering
    \includegraphics[width=1.0\linewidth]{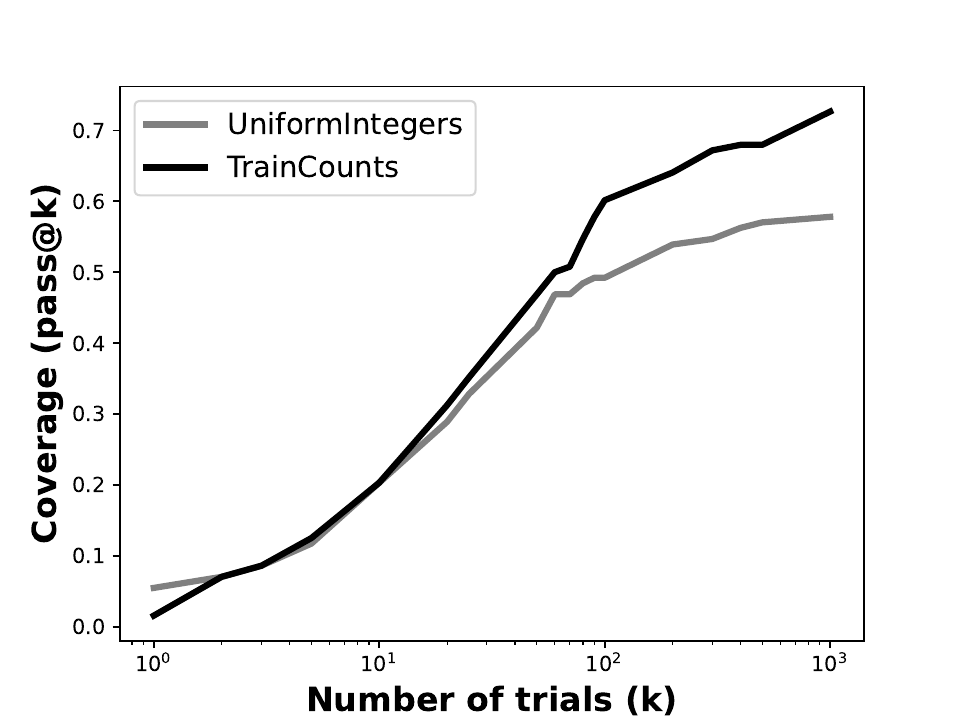}
    \caption{Our guessing strategy (\baseline) vs simply guessing positive integers (\uniformbaseline) on MATH. We see that starting at approximately $k=100$, \baseline{} obtains superior gains compared to \uniformbaseline{}, as it begins guessing some common non-integer answers, or negative integers (see Table \ref{table:math_counts} for qualitative examples). 
    }
    \label{fig:random_baselines_math}
\end{figure}

As mentioned in \S\ref{sec:baselines}, our random baseline is slightly informed, in the sense that it relies on the statistics of the train split of the dataset. To understand how beneficial this is compared to a completely uninformed baseline, we compare \baseline{} to an even simpler enumeration strategy (\uniformbaseline), that simply chooses as $k$ guesses the first $k$ positive integer values. We compare the coverage curves of the two strategies in Figure~\ref{fig:random_baselines_math}.

\section{Additional Results}\label{sec:additional_results}

\subsection{Normalized Coverage}
\begin{figure*}[tb]
\setlength{\belowcaptionskip}{-1pt}
    \centering
    \includegraphics[width=1.0\linewidth]{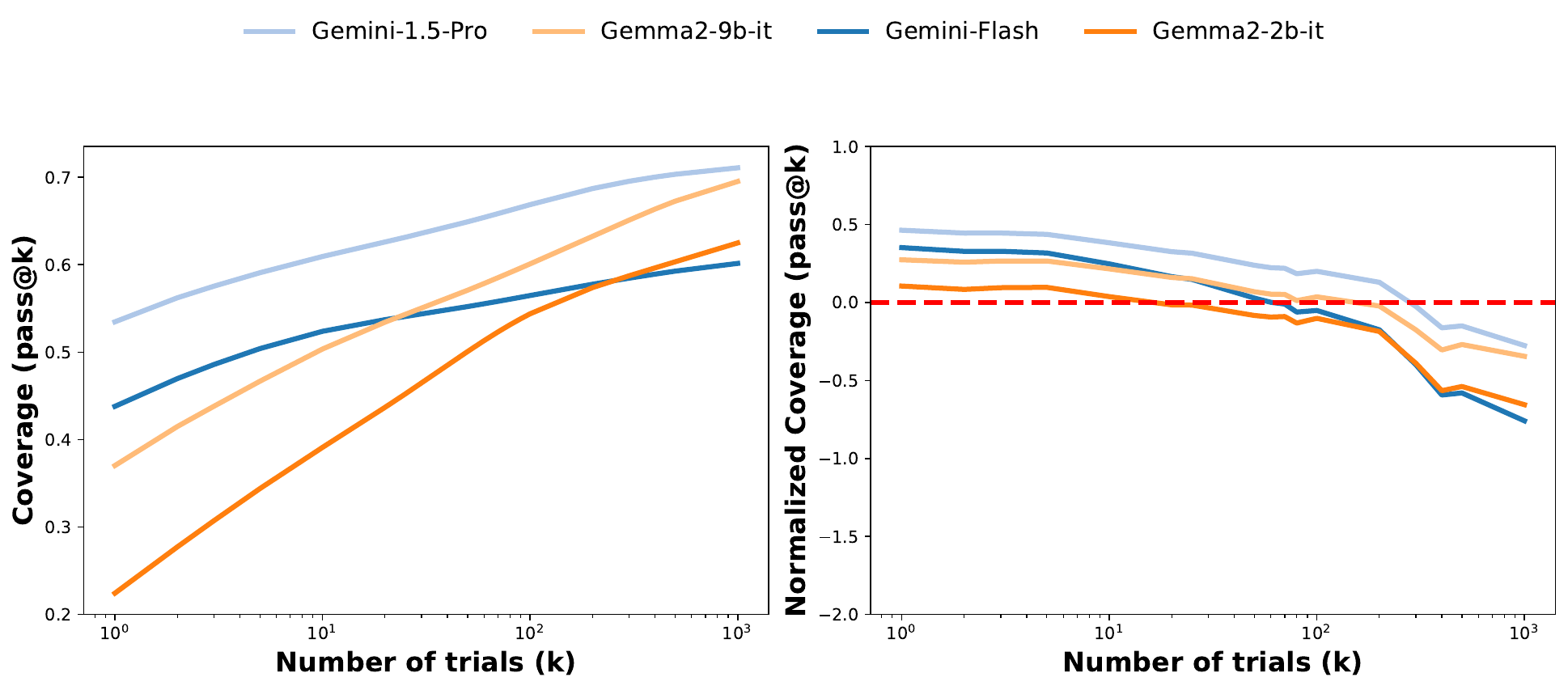}
    \caption{\textbf{Normalized coverage curves for Entity Questions}: We compare standard coverage curves for the models under consideration (LHS) with normalized coverage curves obtained after re-scaling the improvements relative to our \baseline baseline (RHS). While all models demonstrate improved coverage when as $k$ grows, our baseline reveals that for all models, these gains are outperformed by simple answer enumeration as $k$ becomes large.
    }
    \label{fig:normalized_coverage_eq}
\end{figure*}

Figure~\ref{fig:normalized_coverage_eq} shows the normalized coverage for the Entity Questions dataset.

\subsection{Additional \mixture{} Results}
\begin{figure*}[t]
    \centering
    \includegraphics[width=1.0\linewidth]{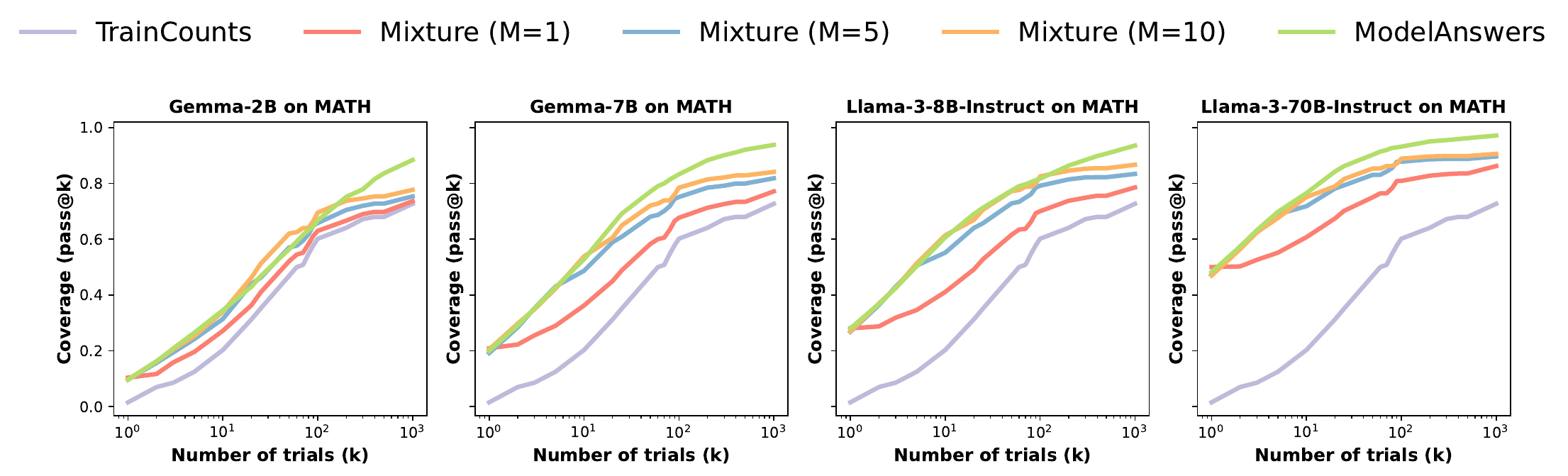}
    \vskip10pt
    \includegraphics[width=1.0\linewidth]{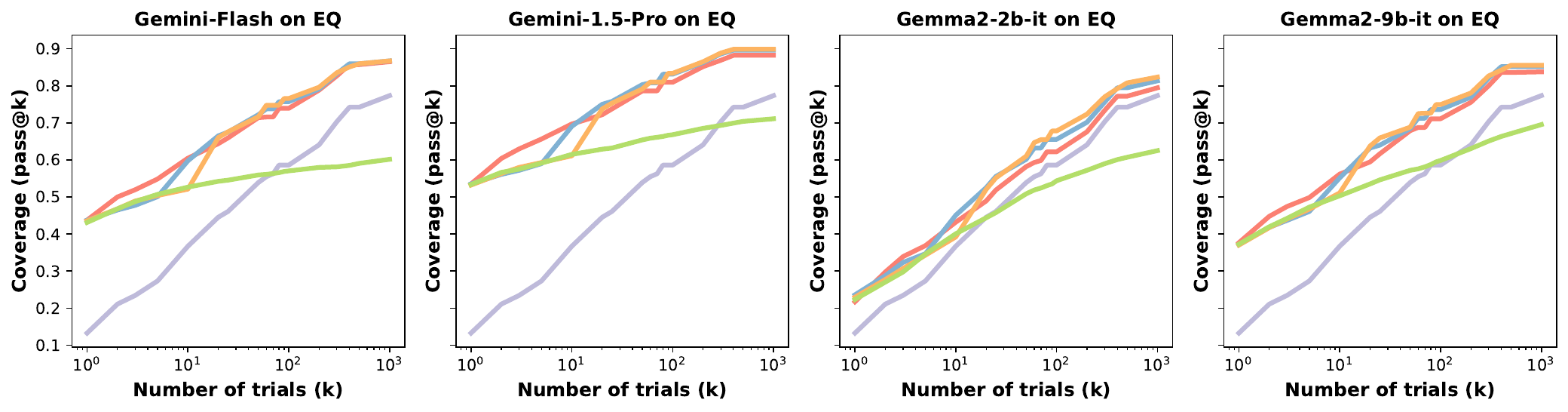}
    \caption{Coverage curves for model sampling and guessing-based answers. \textbf{Top row:} Coverage curves on MATH for the Llama-3 and Gemma models. \textbf{Bottom row:} Coverage curves on EQ for the Gemini and Gemma models. For both MATH and EQ, we calculate \mixture{} for $M$ values of 1, 5 and 10, and compare \mixture{} with \baseline{} and \modelsampling{}.}
    \label{fig:mixture_additional}
\end{figure*}

Figure~\ref{fig:mixture_additional} shows the coverage curves for the four best-performing models (in terms of normalized coverage, see~\ref{sec:results}) and the coverage curves for all four models used for EQ predictions. Add detailed in \S\ref{sec:results}, \mixture{} obtains similar coverage rates as \modelsampling{}.

\subsection{Detailed \mixture{} Results on MATH}
Table~\ref{table:coverage_numbers_math} presents detailed pass@$k$ values for the MATH dataset.
\begin{figure*}[t]
\centering
\begin{tabular}{|l|c|c|c|c|}
\hline
 & \multicolumn{2}{c|}{pass@k} & \% of & \% of \\
\cline{2-3}
 & \modelsampling & \mixture{} & coverage gains & test-time compute \\
\hline
\multicolumn{5}{|l|}{Llama-3-8B-Instruct} \\
\hline
$k=100$ & 0.83 & 0.83 & 100\% & 10\% \\
\hline
$k=1000$ & 0.94 & 0.87 & 92.5\% & 1\% \\
\hline
\multicolumn{5}{|l|}{Llama-3-70B-Instruct} \\
\hline
$k=100$ & 0.93 & 0.90 & 96\% & 10\% \\
\hline
$k=1000$ & 0.97 & 0.91 & 93\% & 1\% \\
\hline
\end{tabular}
\caption{Comparison of \modelsampling vs \mixture for Llama-3 models on the MATH dataset. We see that  combining 10 sampled answers with informed enumeration recovers a significant portion of the coverage gains using only a fraction of the compute budget.  
}
\label{table:coverage_numbers_math}
\end{figure*}

\subsection{Breakdown of the Entity Questions \mixture{} Results}
\label{sec:eq_breakdown}

\begin{figure*}[t]
    \centering
    \includegraphics[width=1.0\linewidth]{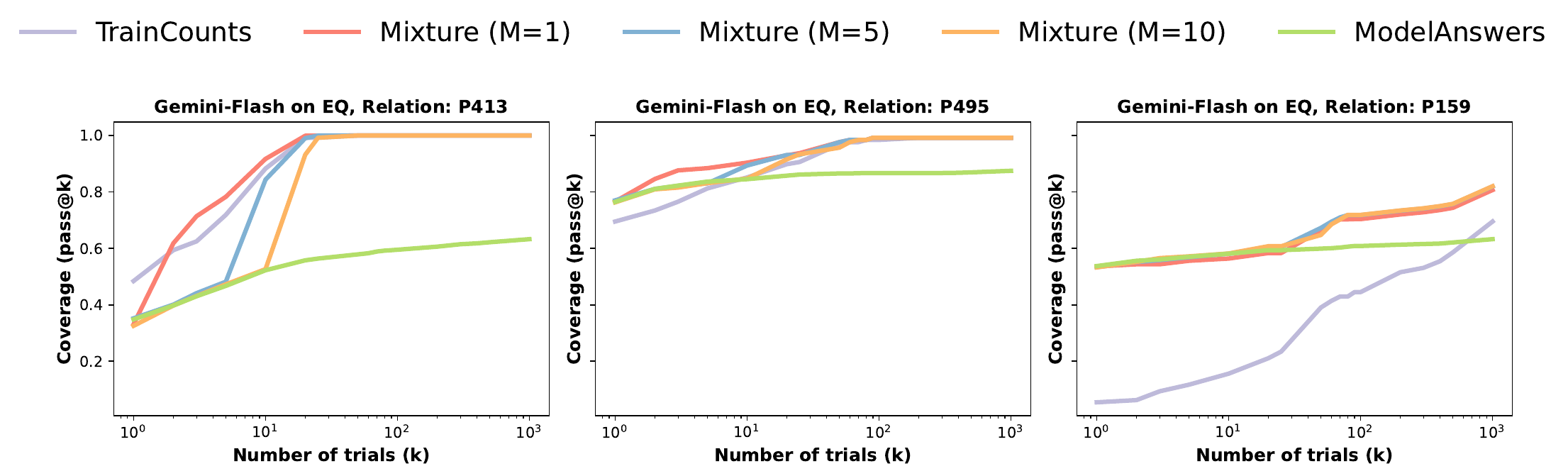}
    \vskip10pt
    \includegraphics[width=1.0\linewidth]{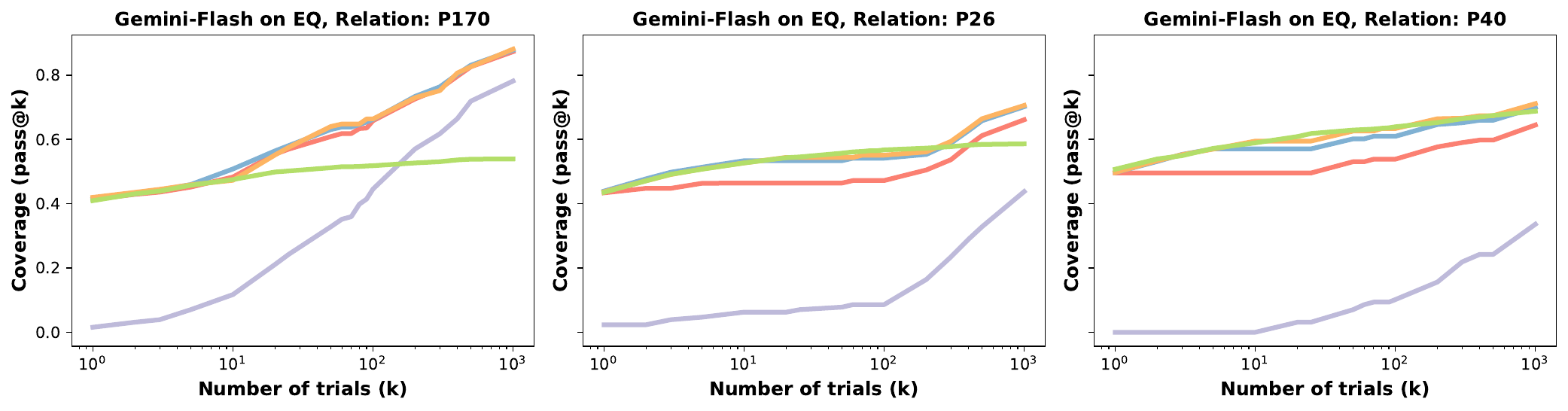}
    \caption{Coverage curves for model sampling and guessing-based answers, divided by relations. P413 and P495 have the smallest set of unique training answers, P26, P40 have the largest set, and P159, P170 are in between. We calculate \mixture{} for $M$ values of 1, 5 and 10, and compare \mixture{} with the \baseline{} and \modelsampling{}.}
    \label{fig:eq_breakdown}
\end{figure*}

For Entity Questions, Figure~\ref{fig:mixture} shows the \mixture{} results for 128 examples sampled from the Entity Questions test set, spanning questions from all of the 24 relations in the dataset. The relations in EQ, however, are of different levels of difficulty. Specifically, when considering \baseline{}, the number of unique train answers vary between relations, ranging from 52 unique answers only (P413) to 9608 (P40). As such, the number of test answers that are included in the training set varies widely. We note that ranking relations according to the number of unique training set answers does not correspond exactly to their ranking according to train-test answer overlap, though the two are strongly connected.

To obtain a clearer image of \baseline{}'s performance for EQ, we select 6 relations: two relations that have the smaller set of training answers (P413, P495), two with the largest set (P26, P40) and two the are ranked in between (P159, P170).\footnote{The full set of relations can be found here: \\\url{https://github.com/princeton-nlp/EntityQuestions}}

As shown in Figure~\ref{fig:eq_breakdown}, results for EQ vary, depending on the tested relation. For the relations with the smallest set of unique training answers, \baseline{} performs better than \modelsampling{}, while for the relations with the largest set of unique  training answers, the opposite is true. For all relations considered, however, \mixture{} usually outperforms \modelsampling{}, even for a small $M$. These results are inline with those presented in Section~\ref{sec:results}.

\end{document}